\documentclass[letterpaper, 10 pt, conference]{formatting/ieeeconf}
\IEEEoverridecommandlockouts                              
\let\labelindent\relax
\usepackage{amsfonts}       

\usepackage{amsthm}
\usepackage{wrapfig}
\usepackage{mathtools}      
\usepackage{amssymb}        
\usepackage{filecontents}
\usepackage{graphicx}       
\usepackage{marginnote}     
\usepackage{marvosym}       
\usepackage{overpic}        
\usepackage{tabularx}
\usepackage{cite}
\usepackage{color}
\usepackage[linesnumbered,algoruled,boxed,lined]{algorithm2e}
\usepackage[normalem]{ulem}
\usepackage{epstopdf}
\usepackage{float}
\usepackage{enumitem}
\usepackage[normalem]{ulem}
\usepackage{lipsum}
\usepackage[a-2b,mathxmp]{pdfx}[2018/12/22]

\newif\ifdraft
\draftfalse

\ifdraft
\usepackage[paperheight=11in,paperwidth=9.5in,
			left=1.25in,right=1.25in,
			top=0.75in,bottom=0.75in,
			heightrounded,marginparwidth=1.2in,
			marginparsep=0.05in]{geometry}
\usepackage{xcolor}
\usepackage{xargs} 
\usepackage[textsize=footnotesize]{todonotes}
\newcommandx{\nt}[2][1=]{\todo[linecolor=red,
			backgroundcolor=red!10,bordercolor=red,#1]{ #2}}
\newcommandx{\jy}[2][1=]{\todo[linecolor=green,
			backgroundcolor=green!10,bordercolor=green,#1]{JY: #2}}
\else
\newcommand{\nt}[1]{{}}
\newcommand{\jy}[1]{{}}
\fi

\newif\iftwocolumn
\twocolumntrue

\newtheorem{proposition}{Proposition}[section]

\theoremstyle{definition}

\theoremstyle{remark}

\SetKwProg{Fn}{Function}{}{}
\SetKwComment{Comment}{$\triangleright$\ }{}

\makeatletter
\def\subsubsection{\@startsection{subsubsection}
                                 {3}
                                 {\z@ \hspace*{1mm}}
                                 {0ex plus 0.1ex minus 0.1ex}
                                 {0ex}
                                 {\normalfont\normalsize\itshape}}
\makeatother

\newcommand{\mpp}{\textsc{MRPP}\xspace}

\newcommand{\cbs}{\textsc{CBS}\xspace}
\newcommand{\ecbs}{\textsc{ECBS}\xspace}
\newcommand{\decbs}{\textsc{DCBS}\xspace}
\newcommand{\secbs}{\textsc{SCBS}\xspace}
\newcommand{\ddm}{\textsc{DDM}\xspace}

\newcommand{\makespan}{\textsc{MKPN}\xspace}
\newcommand{\soc}{\textsc{SOC}\xspace}
\newcommand{\noc}{\textsc{NOC}\xspace}

\title{
Efficient Heuristics for Multi-Robot Path Planning in Crowded Environments
}
\author{Teng Guo   \qquad Jingjin Yu
\thanks{G. Teng, and J. Yu are with the Department of 
Computer Science, Rutgers, the State University of New Jersey, Piscataway, NJ, USA. 
Emails: {\tt\small \{ teng.guo, jingjin.yu\}@rutgers.edu}.
}
\thanks{This work was supported in part by NSF award IIS-1845888 and an Amazon Research Award.
}
}

\begin{document}

\maketitle
\thispagestyle{empty}
\pagestyle{empty}


\begin{abstract}
Optimal Multi-Robot Path Planning (\mpp) has garnered significant attention due to its many applications in domains including warehouse automation, transportation, and swarm robotics. 
Current \mpp solvers can be divided into reduction-based, search-based, and rule-based categories, each with their strengths and limitations. 
Regardless of the methodology, however, the issue of handling dense \mpp instances remains a significant challenge, where existing approaches generally demonstrate a dichotomy regarding solution optimality and efficiency.  
This study seeks to bridge the gap in optimal \mpp resolution for dense, highly-entangled scenarios, with potential applications to high-density storage systems and traffic congestion control. 
Toward that goal, we analyze the behaviors of SOTA \mpp algorithms in dense settings and develop two hybrid algorithms leveraging the strengths of existing SOTA algorithms: \decbs (database-accelerated enhanced conflict-based search) and \secbs (sparsified enhanced conflict-based search). 
Experimental validations demonstrate that \decbs and \secbs deliver a significant reduction in computational time compared to existing bounded-suboptimal methods and improve solution quality compared to existing rule-based methods, achieving a desirable balance between computational efficiency and solution optimality.
As a result, \decbs and \secbs are particularly suitable for quickly computing good-quality solutions for multi-robot routing in dense settings. 

\vspace{2mm}
\noindent Simulation video: \url{https://youtu.be/dZxMPUr7Bqg}\\
\noindent Upon the publication of the manuscript, source code and data will be released at \url{https://github.com/arc-l/dcbs}
\end{abstract}

\section{Introduction}\label{sec:intro}
We study the \emph{labeled} Multi-Robot Motion Planning (\mpp) problem under a graph-theoretic setting, also known as Multi-Agent Path Finding (MAPF). 
The basic objective of \mpp is to find a set of collision-free paths to route multiple robots from a start configuration to a goal configuration.  
In practice, solution optimality is also of key importance; yet optimally solving \mpp in terms of makespan and sum-of-cost is generally NP-hard~\cite{YuLav13AAAI,Sur10,Yu2015IntractabilityPlanar}. 
\mpp algorithms find many important large-scale applications, including, e.g., in warehouse automation for general order fulfillment \cite{wurman2008coordinating}, grocery order fulfillment \cite{mason2019developing}, and parcel sorting \cite{wan2018lifelong}.
Other application scenarios include formation reconfiguration~\cite{PodSuk04}, agriculture~\cite{cheein2013agricultural}, object 
transportation~\cite{RusDonJen95}, swarm robotics \cite{preiss2017crazyswarm}.

Given the potential of employing its solutions in a wide range of impactful applications, even though \mpp had been studied since the 1980s in the robotics domain~\cite{KorMilSpi84,ErdLoz86,LavHut98b,GuoPar02}, it remains a highly active research topic. 
Many effective algorithms, for example~\cite{YuLav16TRO, boyarski2015icbs, cohen2016improved}, have been proposed recently that balance fairly well between computational efficiency and solution optimality.
Existing \mpp algorithms have been tested on randomly generated instances and yield decent performance for instances with relatively limited robot-robot interactions, i.e., either the number of robots is limited, or the density of robots is relatively low.
However, they frequently fail in instances that are both large and dense.
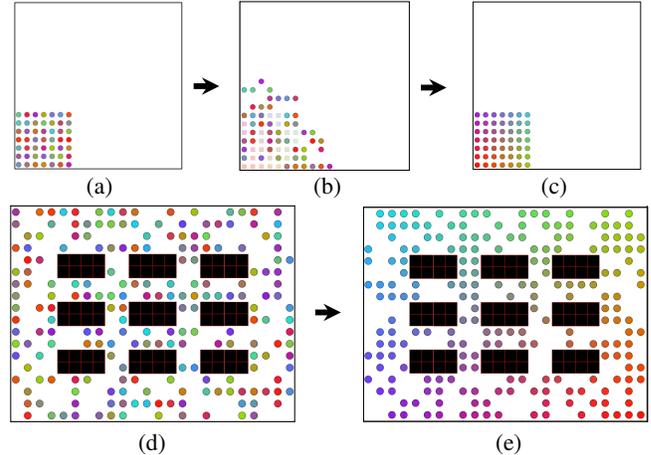
\begin{figure}[t]
\vspace{2mm}
    \centering
  \begin{overpic}               
        [width=1\linewidth]{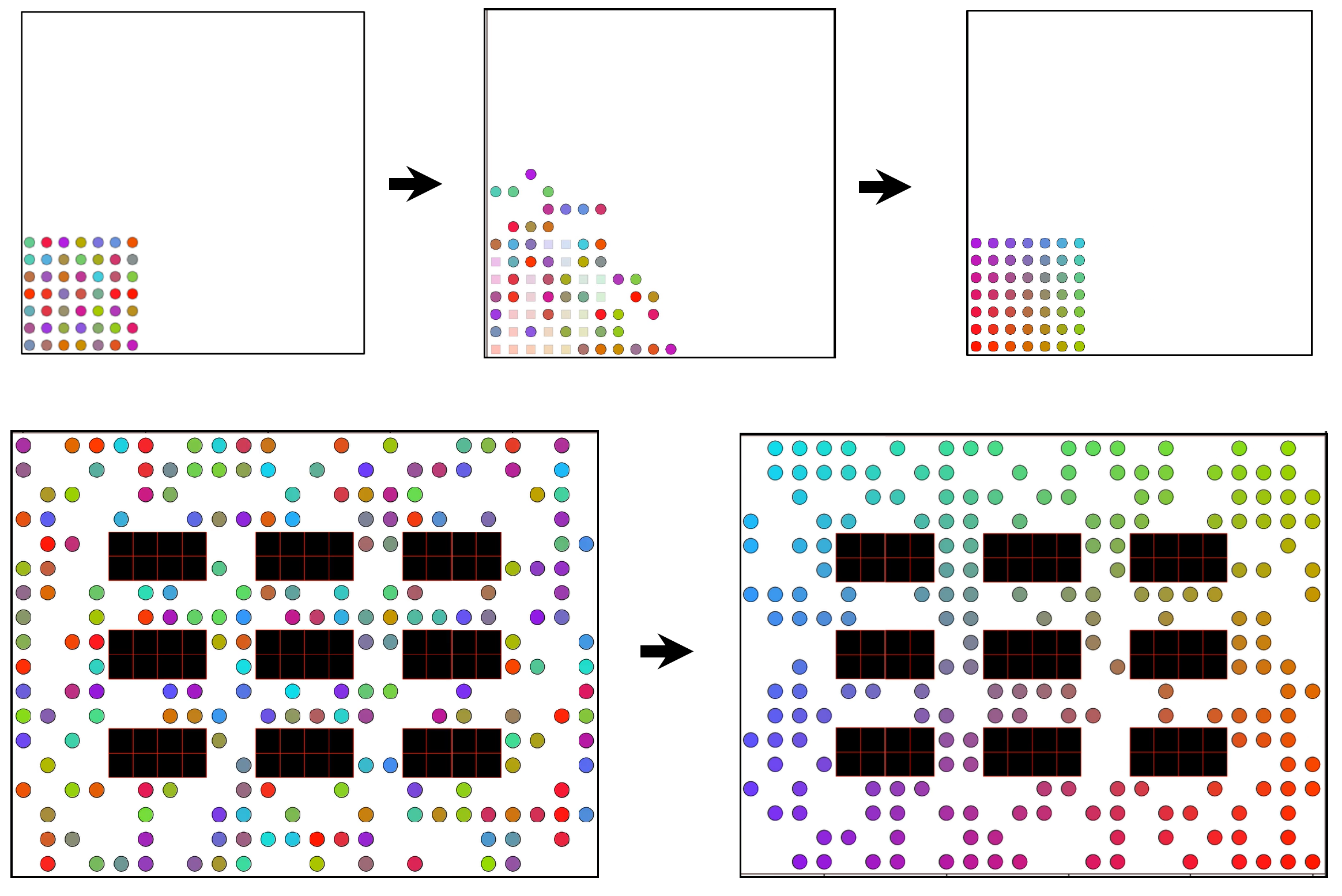}
             \small
             \put(12.5, 36.5) {(a)}
             \put(47.5,36.5) {(b)}
             \put(82.5, 36.5) {(c)}
             \put(20.5, -3) {(d)}
             \put(75.5, -3) {(e)}
        \end{overpic}
\vspace{-3mm}
    \caption{(a)-(c) A challenging \emph{locally-dense} \mpp example on $20\times 20$ map with 49 robots. It requires rearranging the robots from the start configuration (a) to the goal configuration (c). By ``sparsifying'' the configuration using our methods, as shown in an intermediate step (b), the problem can be solved quickly with decent solution optimality. (d)-(e) A challenging \emph{globally-dense} \mpp example on a $24\times 18$ warehouse map with 203 robots. In both settings, each robot has a unique start and goal.}
    \label{fig:dense_example}
\end{figure}%
Recently, \mpp algorithms have been applied in high-density applications, such as autonomous vehicle parking systems \cite{Guo2023TowardEP,okoso2022high}, to increase space utilization efficiency.
In such dense scenarios, robots' motions are strongly correlated and may block the paths of each other, which makes the problem extremely difficult for existing \mpp solvers.

\textbf{Results and contributions.}
This research proposes efficient heuristics and uses them to build complete solvers for tackling dense and difficult \mpp instances. 
We address two classes of dense \mpp: \emph{globally} dense instances where the number of robots is large with high average robot density (more than $40\%$, see Fig.~\ref{fig:dense_example}(c)), and \emph{locally} dense instances where the robot distribution is unbalanced with high local robot density (i.e. $100\%$, see Fig.~\ref{fig:dense_example}(a)-(b)).

We develop two hybrid \mpp algorithms to address the above-mentioned challenges.
%
%
%
%
%
%
In the first algorithm, we introduce a (motion-primitive) database-based conflict resolution mechanism inspired by \cite{han2019ddm} to augment a conflict-based search \cite{sharon2015conflict}.
%
%
%
We also design a set of rules to maintain the solution quality as well as the completeness of the resulting algorithm. 
We call the algorithm \decbs, standing for \emph{database-accelerated enhanced conflict-based search}.  

While our first algorithm works for both globally dense and locally dense scenarios, our second algorithm is designed specifically for locally dense instances.
Inspired by \cite{guo2022sub}, we first convert the challenging configuration to a \emph{sparsified} configuration, which is relatively easier to solve, using \emph{unlabeled} \mpp planning solutions. 
To reduce the extra overhead of the conversion, we adopt a best-first heuristic for finding a proper sparsified configuration and a path refinement technique for better concatenating the intermediate paths.
We call the second algorithm \secbs, standing for \emph{sparsified enhanced conflict-based search}.  

Experiments on diverse environment maps demonstrate the effectiveness of our proposed methods in solving instances with robot densities greater than $60\%$-$70\%$ with a high success rate and decent levels of solution quality. 
\decbs and \secbs outperform previous \mpp algorithms in terms of combined speed and solution quality.

\textbf{Organization.}
The rest of the paper is organized as follows. Sec.~\ref{sec:problem} covers the preliminaries, including the problem formulation and two suboptimal algorithms \ecbs and \ddm. 
In Sec.~\ref{sec:algo-trans}-Sec.~\ref{sec:algo-ddecbs}, we describe our heuristics and algorithms for solving dense \mpp. 
We perform thorough evaluations and discussions of the proposed algorithms in Sec.~\ref{sec:evaluation} and conclude with Sec.~\ref{sec:conclusion}.

%

\section{Related Research} 
\mpp/MAPF has been widely studied in the field of robotics. 
In the static or one-shot setting \cite{stern2019multi},  given a graph environment and a number of robots with each robot having a unique start position and a goal position,  the task is to find collision-free paths for all the robots from start to goal.
It has been proven that solving one-shot \mpp optimally in terms of minimizing either makespan or sum of costs is NP-hard \cite{surynek2010optimization,yu2013structure}.
Moreover, it is also NP-hard to approximate within any constant factor less than 4/3 if the solution makespan is to be minimized~\cite{ma2016multi}.

Existing solvers for \mpp can be broadly categorized into \emph{reduction-based}, \emph{search-based}, and \emph{rule-based}.

Reduction-based solvers reduce \mpp to other well-studied problems, such as ILP\cite{yu2016optimal}, SAT\cite{surynek2010optimization} and ASP\cite{erdem2013general}.
These solvers are able to find optimal solutions and are efficient for small and dense instances.
There are dividing-and-conquer heuristics for enhancing their scalability \cite{guo2021spatial,yu2016optimal} at the cost of optimality.
Unfortunately, these reduction-based methods are still incapable of dealing with the extremely dense scenarios we study in this paper.

%
Another more popular approach develops search-based algorithms for \mpp problems that can be viewed are high-sophisticated A* \cite{hart1968formal} variants. 
Coupled A* \cite{silver2005cooperative}, ICTS \cite{sharon2013increasing}, and CBS \cite{sharon2015conflict} are optimal solvers and efficient on large maps but with sparse robots.
ECBS \cite{barer2014suboptimal} is the bounded-suboptimal version of CBS with enhanced scalability and bounded suboptimality.
The search-based algorithms rely heavily on the heuristic for reducing the number of node expansions, i.e., Number Of Conflicts (\noc) which is used in \ecbs. 
In dense scenarios, since robots are strongly correlated, this heuristic is not enough since there are lots of nodes with the same \noc.
As a result, the number of nodes needed to expand to find a solution for existing \cbs variants grows exponentially with respect to the robot density, even if there is only a small number of robots.

Rule-based solvers are another class of suboptimal \mpp solvers.
Prioritized planners \cite{silver2005cooperative, okumura2019priority} assign priorities to robots and low-priority robots avoid conflicts with high-priority robots by treating them as dynamic obstacles.
This is efficient but can easily cause dead-lock issues in dense scenarios, which leads to a low success rate.
Another class of rule-based solvers introduces motion primitives for swapping the position of robots, such as Push-And-Swap \cite{luna2011push} and Rubik Table \cite{szegedy2020rearrangement,guo2022sub,GuoFenYu22IROS}.
They are polynomial-time algorithms and can even solve extremely dense instances, but the solution quality is far from optimal.
DDM\cite{han2019ddm} resolves the inter-robot conflicts efficiently by utilizing the precomputed motion primitive within the $3\times 3$ sub-grid.
It finds near-optimal solutions when the robot density is not high but has the same optimality issue in dense scenarios.

\section{Preliminaries}\label{sec:problem}
\subsection{Multi-Robot Path Planning on Graphs}
A graph-based Multi-Robot Path Planning (\mpp) problem is defined on a graph $\mathcal{G} = (\mathcal{V},\mathcal{ E})$. 
We assume that $\mathcal{G}$ is a grid graph. 
That is, given integers $w$ and $h$ as the graph's {\em width} and {\em height}, 
the vertex set can be represented as $\mathcal{V} \subseteq \{(i, j) \mid 1 \leq i \leq w, 1 \leq j \leq h,i\in \mathbb{Z},j\in \mathbb{Z}\}$. 
The graph is $4$-way connected, i.e., for a vertex $v = (i, j)$, the set of its neighboring 
vertices are defined as $\mathcal{N}(v) = \{(i + 1, j),(i - 1, j),(i, j + 1),(i, j - 1)\} \bigcap \mathcal{V}$. 
The problem involves $n$ robots $r_1, \dots, r_n$, 
where each robot $r_i$ has a unique start state $s_i \in \mathcal{V}$ and a unique goal state $g_i \in \mathcal{V}$. 
We denote the joint start configuration as $X_S = \{s_1, \dots, s_n\}$ and the goal configuration as $X_G = \{g_1, \dots, g_n\}$. 

The objective of \mpp is to find a set of feasible paths for all robots. 
Here, a {\em path} for robot $r_i$ is defined as a sequence of $T + 1$ vertices 
$P_i = (p_i^0, \dots, p_i^T)$ that satisfies: 
(i) $p_i^0 = s_i$; 
(ii) $p_i^T = g_i$; 
(iii) $\forall 1 \leq t \leq T, p_i^{t - 1} \in N(p_i^t)$. 
Apart from the feasibility of each individual path, for $P$ to be collision(conflict)-free , 
$\forall 1 \leq t \leq T, 1 \leq i < j \leq n$, $P_i, P_j$ must satisfy 
\begin{enumerate}[leftmargin=7mm]
    \item There is no \emph{vertex collision}: $p_i^t \neq p_j^t$; 
    \item There is no \emph{edge collision}: $ (p_i^{t - 1}, p_i^t) \neq (p_j^t, p_j^{t - 1})$.
\end{enumerate}
and the following criteria are used to evaluate solution quality:
\begin{enumerate}[leftmargin=7mm]
    \item Makespan (\makespan): the time required to move all robots to their desired positions;
    \item Sum-of-cost (\soc): the cumulative cost function that sums over 
all robots of the number of time steps required to reach the goals. For each robot, denoting $t_i$ 
such that $\forall t_i \leq t \leq T, p_i^t = g_i$, the sum-of-costs objective is calculated as 
$\min \sum_{1 \leq i \leq n} t_i$. 
\end{enumerate}
In general, these two objectives create  a  Pareto  front \cite{yu2013structure}, and it  is  not always possible to simultaneously optimize these objectives.

\subsection{Enhanced Conflict Based Search (ECBS)}

\ecbs($w_1$)\cite{barer2014suboptimal} is a variant of CBS \cite{sharon2015conflict} that is  $w_1$-suboptimal, which employs the focal search method \cite{pearl1982studies}  in both its high-level and low-level searches  rather than best-first searches. 

A focal search, like A*, uses an OPEN list whose nodes $n$ are sorted in increasing order of their $f$-values $f(n)=g(n)+h(n)$, where $h(n)$ are the primary heuristic values. 
Unlike A*, a focal search with suboptimality factor $w_1$ also uses a FOCAL list of all nodes currently in the OPEN list whose $f$-values are no larger than $w_1$ times the currently smallest $f$-value in the OPEN list.
The nodes in the FOCAL list are sorted in increasing order of their secondary heuristic values.
A* expands a node in the OPEN list with the smallest $f$-value, but a focal search expands a node in the FOCAL list instead with the smallest secondary heuristic value.
Thus, the secondary heuristic values should favor a node in the FOCAL list close to a goal node to speed up the search and thus exploit the leeway afforded by $w_1$ that A* does not have available. 
If the primary heuristic values are admissible, then a focal search is $w_1$-suboptimal.
The secondary heuristic values can be inadmissible.

The high-level and low-level searches of ECBS($w_1$) are both focal searches.
During the generation of a high-level node $N$, ECBS($w_1$) performs a low-level focal search with OPEN list $\text{OPEN}_i(N)$ 
and FOCAL list $\text{FOCAL}_i(N)$ for the robot $i$ affected by the added constraint. 
The number of collisions(\noc) is used as the secondary heuristic value for the high-level and low-level searches, allowing ECBS ($w_1$) to generate high-level nodes with fewer collisions compared to CBS, which improves its efficiency. 
However, the path costs can become large for ECBS($w_1$) with large values of $w_1$ due to the larger leeway afforded by $w_1$. 
The robots might move around in wiggly lines, increasing the chance of collisions, thus increasing the number of collisions in the high-level and low-level nodes of ECBS($w_1$) and slowing it down.
Thus, larger values of $w_1$ do not necessarily entail smaller runtimes of ECBS($w_1$).
In this paper, the \soc suboptimality bound is chosen to be $w_1=1.5$, which is a good choice according to the original paper\cite{barer2014suboptimal}.

\subsection{DDM}
\ddm~\cite{barer2014suboptimal}, standing for \emph{\textbf{d}iversified path and \textbf{d}atabase-driven \textbf{m}ulti-robot path planner}, is a fast suboptimal \mpp solver.
It first generates a shortest  path  between  each  pair  of  start  and  goal  vertices and then resolves local conflicts among the initial paths. 
In generating the initial paths, a path diversification heuristic is introduced that attempts to make the path ensemble use all  graph  vertices  in  a  balanced  manner,  
which  minimizes the  chance  that  many  robots  aggregate  in  certain  local areas, causing unwanted congestion, in order to reduce the number of conflicts of the initial paths. 

Then, in resolving the path conflicts, a database resolution heuristic is introduced, which builds a min-makespan solution database for  all $2\times 3$ and $3\times3$ sub-problems and  ensures  quick  local  conflict  resolution  via  database retrievals. 
Specifically, for each conflicting robot pair in each step, \ddm tries to find a $2\times 3$ or $3\times 3$ subgraph that contains these robots.
Temporary goals are assigned to the robots within the subgraph to resolve the conflict.
The paths for routing them to the temporary goals can be obtained easily by accessing the precomputed database.
Obviously, each time resolving a conflict using subgraphs will introduce an extra overhead to the paths' length.
In dense environments, the number of conflicts needed to resolve is high, and as a result, \ddm can be very suboptimal under these scenarios.

\section{Database Conflict Resolution in \ecbs}\label{sec:algo-ddecbs}
\begin{figure}[!htbp]
    \centering
    \includegraphics[width=\linewidth]{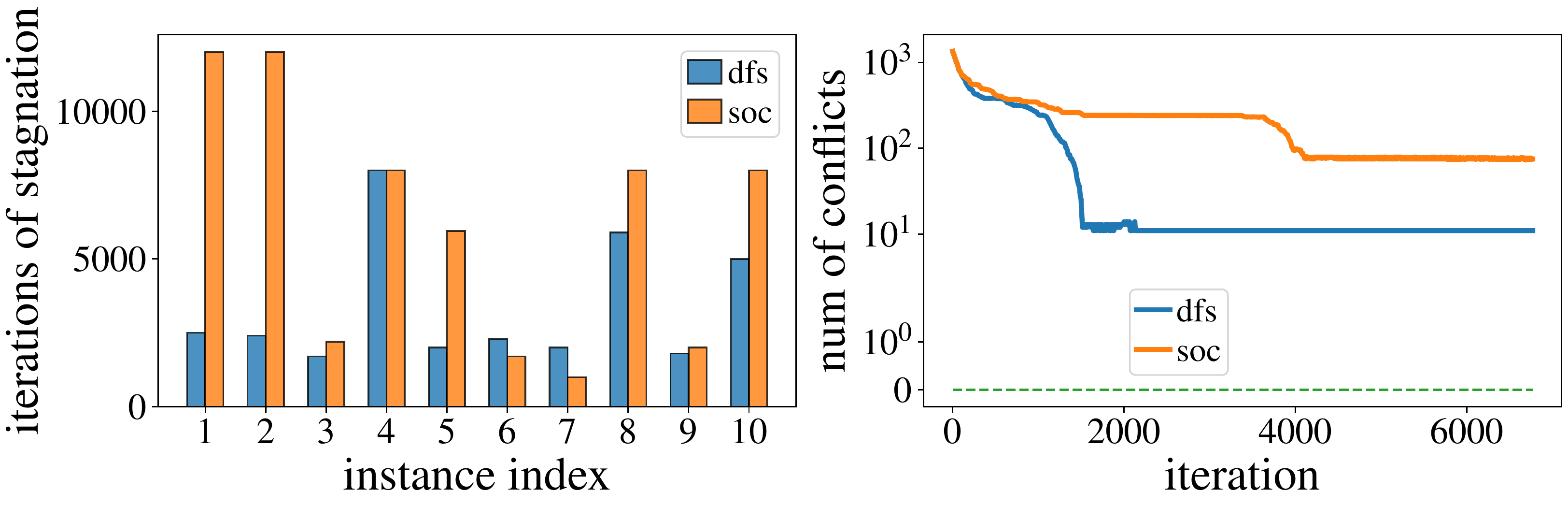}
    \caption{Left:  Number of iterations to enter \noc stagnation on 10 random instances on $20\times 20$ map with 272 robots. Right: An example of \noc stagnation phenomenon when applying \ecbs to solve a dense instance in $20\times 20$ map with 272 robots.}
    \label{fig:stuck_example}
    \vspace{2mm}
\end{figure}

Dense instances are challenging for \ecbs to solve.
Fig.~\ref{fig:stuck_example} shows an example of applying \ecbs to solve a dense instance in $20\times 20$ that has 272 robots with random starts and goals.
The \noc decreases as the number of iterations of high-level expansion increases.

When robot density is not very high, every time a constraint is added to a high-level node, it will lead to the \noc decreasing by at least one.
Meanwhile, the \noc of the initial node is not very large for sparse instances.
Therefore, \ecbs finds a conflict-free solution efficiently when robot density is not very high.
However, when robot density is high, the \noc will be stuck at some non-zero point.
This is because robots' interactions are strongly correlated in high-density settings.
Adding one constraint to resolve  a given conflict may cause the low-level planner to find a path conflicting with another robot.
As a result, the \noc does not decrease and there would be a large number of nodes with the same \noc in the OPEN list.
The stagnation of \noc will continue for a long period of high-level expansion until it accidentally expands the correct node. 
Even worse, it is possible that \ecbs cannot find a feasible solution after expanding all the nodes with the stagnated \noc in the current OPEN list  and it needs to expand nodes with higher \noc,
which makes \ecbs very inefficient.

To address the issue, we propose \emph{database-accelerated enhanced conflict-based search} (\decbs) (Alg.~\ref{alg:decbs}), which introduces a database-driven conflict resolution mechanism into \ecbs to speed up the high-level expansion and circumvent the \noc stagnation.
\decbs expands the high-level nodes regularly as \ecbs does initially.
When the \noc of the node to expand drops to a specific point, the database conflict resolution mechanism is triggered and is applied to that node (Line 7).
The paths of the current node are used as the initial paths for conflict resolution.
We apply the database heuristics to resolve all the conflicts in the paths, in a local $2\times 3$ sub-graph or $3\times 3$ sub-graph. 
There is the possibility that we cannot find a sub-graph for a pair of conflicting robots if the map is not a low-resolution graph \cite{han2019ddm}.

When we could not resolve the conflicts, we return to the \ecbs high-level expansion routine and continue to use focal search in the low level to resolve the conflicts.
If $\texttt{DbResolution}$ succeeds in finding a solution, to ensure the solution quality, we check if the \makespan(\soc) suboptimality ratio of paths is within the bound of $w_2$, where $w_2>w_1$ is another user-defined suboptimality bound. 
When the solution, after resolving all the conflicts using database heuristics satisfies the optimality need, we return the solution.
Otherwise, we continue the \ecbs high-level expansion.
\begin{algorithm}
\begin{small}
\DontPrintSemicolon
\SetKwProg{Fn}{Function}{:}{}
\SetKwFunction{Fcbs}{\decbs}
\SetKw{Continue}{continue}
 \caption{\decbs Outline\label{alg:decbs}}
$\text{Root}\leftarrow$\texttt{InitializeRoot()}\;
\vspace{1mm}
$\text{OPEN}.push(\text{Root})$\;
\vspace{1mm}
\While{$\textsc{OPEN}\neq \emptyset$}{
\vspace{1mm}
$\text{FOCAL}\leftarrow \text{PriorityQueue}(\{n\in \text{OPEN}|n.\soc<\omega_1\cdot n.LB\})$\;
\vspace{1mm}
$N\leftarrow \text{FOCAL}.pop()$\;
\vspace{1mm}
$\text{OPEN}.remove(N)$\;
{\color{blue}{
\vspace{1mm}
\If{$\texttt{DbTriggered}(N)=true$}{
$success\leftarrow \texttt{DbResolution}(N)$\;
\If{$success= true$ and \texttt{CheckOptimality}($N,\omega_2$)=true}{
\Return $N.paths$\;
}
}
}
}
\vspace{1mm}
$conflict\leftarrow\texttt{FindFirstConflict}(N)$\;
\vspace{1mm}
\For{$r$ involved in $conflict$} {
$N'\leftarrow N.copy()$\;
$C'\leftarrow \texttt{ResolveConflict}(conflict,r)$\;
$N'.constraints.add(C')$\;
$success\leftarrow\texttt{LowLevelPlanner}(N')$\;
\If{success=true}{$\text{OPEN}.push(N')$}\;
}
}
\end{small}
\end{algorithm}

Because \decbs preserves the general structure of \ecbs, the bounded-suboptimality guarantee of \ecbs is inherited.
\begin{proposition}
\decbs is complete and $w_2$ bounded-suboptimal.
\end{proposition}

To make \decbs efficient, we observe that we must pay careful attention to a few key points.
First, we must choose the right time to trigger the database-driven conflict resolution. 
Second, the \noc of the node should drop as quickly as possible and enter the \noc stagnation state as fast as possible.
For example, in Fig.~\ref{fig:stuck_example}, the blue curve is better than the yellow curve for \decbs since it ``converges" to the stagnation point in a much shorter time.
Third, if we want a suboptimality guarantee at some desired level, $\omega_2$ should be also carefully chosen to balance runtime and optimality. 
We certainly hope that the \noc of the node to apply database conflict resolution is small enough.
Otherwise, if the node still contains a lot of conflicts, the resulting paths would be very sub-optimal.
On the other hand, in dense scenarios, if the desired \noc is too small,  it might take a very long time for the \noc of the node to drop to this value.

Based on the observations above, we introduce several additional techniques to enhance the performance of \decbs.
We first apply a DFS-like expansion mechanism to speed up the \noc descent.
The high level  is a best-first search which always first expands the node with the smallest \noc in the FOCAL.
When the density is high, as mentioned before, adding one constraint for avoiding a given conflict may cause a new conflict in the child node. 
As a result, there would be a lot of nodes with the same \noc.
The high-level may randomly pick one node among them, which can be very inefficient. 
Using \soc of the paths as the tie-breaker is a common way for the high-level search.
However, this makes the high-level search inclined to expand nodes with shorter paths, which is efficient in sparse environments.
In dense environments, robots inevitably need to take more detours, and shorter paths do not really have fewer conflicts.

Since shorter paths can be wasteful to sift through, we speed up the expansion in \decbs by adopting a DFS-like strategy. 
Specifically, among the nodes with the same \noc, we choose to first explore the node that was \emph{most lately} pushed to the OPEN list.
With this choice, the high-level search is more inclined to explore as far as possible along a branch.
As it goes deeper along a branch more quickly, the \noc descent enters stagnation in less time. 
In the example from Fig.~\ref{fig:stuck_example}, the blue curve uses the second strategy while the orange one uses \soc as the tie-breaker.
Using DFS-like expansion strategy leads to ``steeper" \noc descent, which is more suitable for \decbs.

In our method, the proper time to trigger the database can be based on the following rules:
\begin{enumerate}
    \item The \noc of the current high-level node is less than a predefined value $NOC_p$.
    \item The \noc is in stagnation. For example, the value-change of the \noc in the high-level expansion is within a range for a number of iterations.
\end{enumerate}

Rule (1) is straightforward. 
The solution quality of the database conflict resolution mechanism is heavily affected by the \noc of the node. 
If the \noc of the current node is small enough, applying the database to resolve the conflicts will introduce only small overheads, and leads to a solution with good quality.
However, the suitable $NOC_p$ may vary in different maps and densities.
If the $NOC_p$ is set very small in a very dense environment, the high-level search may enter \noc stagnation before its \noc drops below $NOC_p$.
As a result, it takes a long time to trigger the database conflict resolution.
%
%
%
In rule (2), the database conflict resolution  is applied when the searching enters \noc stagnation, which is more flexible than the rule (1).
The main drawback of this rule is that there might be multiple stagnations.
If the high-level search enters one stagnation but the \noc is still large, the final solution can be very sub-optimal.


\section{Configuration Sparsification}\label{sec:algo-trans}
In this section, we describe \emph{sparsified enhanced conflict-based search} (\secbs), a new algorithm for solving the locally-dense \mpp instances. 
In locally-dense \mpp instances, the total number of robots in a map is not necessarily high. 
But in the start/goal configurations, robots might be distributed unevenly.
In these instances, the local density at some locations is extremely high, i.e., $\approx 100\%$.
Assume that the local area of the vertex $v$ is the $W\times W$ square area centered at $v$.
The local density at vertex $v$ is defined as $\rho_l(v)=\frac{n_v}{A_v}$, where $n_v$ is the number of robots located in the local area of $v$ and $A_v$ is the number of non-obstacle vertices in the local area of $v$.

%
%
The hybrid \secbs algorithm is outlined in Alg.~\ref{alg:secbs} and Alg.~\ref{alg:greedy}.
The basic idea of \secbs is to convert the congested configurations into some intermediate configurations that are less dense and correlated and thus easier to solve. 
\secbs first tries to find an intermediate start configuration $X_S'$ and an intermediate goal configuration $X_G'$ which are more sparse than original starts and goals.
Then the original problem breaks into three sub-problems, $P_1(\mathcal{G},X_S,X_S'),P_2(\mathcal{G},X_S',X_G'),P_3(\mathcal{G},X_G',X_G)$.
Since the intermediate states are less dense than the original starts and goals, robots are less correlated, and as a consequence, solving $P_2(\mathcal{G},X_S',X_G')$ using \ecbs takes less time than solving the original problem.
While for $P_1$ and $P_3$, they can be formulated as unlabeled \mpp and be solved in polynomial time using  algorithms in \cite{yu2012distance,yu2013multi} (line 4-5). 
The final solution can be obtained by merging the paths for the sub-problems (line 7).

Obviously, the sparsification procedure introduces additional overhead on the optimality.
Finding a good intermediate state is essential for balancing the computation time and solution quality.
The intermediate configurations should try to satisfy the following:
(i). $X_S'$ and $X_G'$ should be close to the original states as much as possible; 
(ii). The local density for each robot is controlled under a preferred robot density $\rho^{*}$, if possible.
%
%
Finding the intermediate state can be formulated as an optimal assignment problem, which may be solved using integer linear programming.
However, this would be very time-consuming.
Instead, we develop an efficient suboptimal greedy algorithm for finding the assignment.

\begin{algorithm}
\begin{small}
\DontPrintSemicolon
\SetKwProg{Fn}{Function}{:}{}
\SetKw{Continue}{continue}
  \KwIn{Starts $X_S$, goals $X_G$, preferred density $\rho^{*}$}
  \Fn{\textsc{SECBS}({$S,G$})}{
 \caption{\secbs \label{alg:secbs}}
\vspace{1mm}
$X_S''\leftarrow\texttt{SparsifyConfig}(X_S,X_G,\rho^{*})$\;
\vspace{1mm}
$X_G''\leftarrow\texttt{SparsifyConfig}(X_G,X_S,\rho^{*})$\;
\vspace{1mm}
$X_S',P_S\leftarrow \texttt{UMRPP}(X_S,X_S'')$\;
\vspace{1mm}
$X_G',P_G\leftarrow \texttt{UMRPP}(X_G,X_G'')$\;
\vspace{1mm}
$P_M\leftarrow \texttt{ECBS}(X_S',X_G')$\;
\vspace{1mm}
$solution\leftarrow \texttt{Merge}(P_S,P_M,P_G)$\;
\vspace{1mm}
\Return $solution$\;
}
\end{small}
\end{algorithm}

Alg.~\ref{alg:secbs} describes how we find the intermediate configuration.
It runs in a decoupled manner and finds the best location for each robot one  by one greedily.
For each robot $i$, we use A* to  explore the nodes in the graph where the A* heuristic is set to be the sum of the distance  from its start and goal.
For the node $u$ to expand, we check if we choose $u$ as the intermediate vertex for robot $i$ whether the local density at each vertex in CONFIG  is still less than $\rho^{*}$.
If that is true, we set $u$ as an intermediate vertex and add it to CONFIG.
The configurations found by the greedy algorithm are used as the unlabeled configurations $X_S''$ and $X_G''$.
The unlabeled \mpp solver finds the intermediate paths $P_S$ and $P_G$ and assigns the intermediate vertices to the robots to get the labeled configurations $X_S', X_G'$.
\begin{algorithm}
\begin{small}
\DontPrintSemicolon
\SetKwProg{Fn}{Function}{:}{}
\SetKw{Continue}{continue}
 \caption{SparsifyConfig \label{alg:greedy}}
 $\text{CONFIG}\leftarrow\{\}$\;
\vspace{1mm}
\For{$i$ in $[1,...,n]$}{
$n\leftarrow (s_i,dist(s_i,g_i))$\;
\vspace{1mm}
$\text{OPEN}\leftarrow \{n\}$\;
\vspace{1mm}
$\text{CLOSE}\leftarrow \{\}$\;
\vspace{1mm}
\While{$\textsc{OPEN}\neq \emptyset$}{
\vspace{1mm}
$(u,f)\leftarrow \text{OPEN}.pop()$\;
\vspace{1mm}
\If{$u\in \textsc{CLOSE}$}{continue\;}
\vspace{1mm}
$\text{CLOSE}.add(u)$\;
\vspace{1mm}
\If{$\texttt{CheckDensity}(u,\textsc{CONFIG},\rho^{*}) \vspace{1mm}
\wedge u\not\in \textsc{CONFIG}$}{
\vspace{1mm}
$\text{CONFIG}.add(u)$\;
\vspace{1mm}
break\;
}
\For{$v\in u.neighbors$}{
\vspace{1mm}
$f\leftarrow dist(v,s_i)+dist(v,g_i)$\;
\vspace{1mm}
$\text{OPEN}.push((v,f))$\;
}
}
}
\Return CONFIG\;
\end{small}
\end{algorithm}

As for merging the paths, simply concatenating the paths which may make the solution very suboptimal in terms of \soc \cite{guo2021spatial}.
This is because robots need to be synchronized to execute the planned paths of each subproblem and some of the robots have to wait unnecessarily. 
We use the method based on Minimum Communication Policy (MCP)~\cite{ma2016information} in \cite{Guo2023TowardEP}. 
This method tries to move the robots to their next vertex in their original plan as quickly as possible, which leads to a solution with better \soc optimality.

\section{Evaluation}\label{sec:evaluation}
In  this  section,  we  evaluate the  proposed algorithms on dense instances.
All experiments are performed on an Intel\textsuperscript{\textregistered} Core\textsuperscript{TM} i7-9700 CPU at 3.0GHz.
We compare the proposed methods with \ecbs($w=1.5$)\cite{barer2014suboptimal} and \ddm~\cite{han2019ddm}.
%
All algorithms are implemented in C++.
We evaluate the makespan, \soc, computation time, and success rate on a diverse set of maps and under different robot density levels.
We repeated each experiment 20 times for each specific setting using different randomly generated instances for the agents, and report the mean values.
Each algorithm is given 60 seconds time limit for each instance and the success rate is the number of solved instances divided by the total number of instances.  
%
The source code and evaluation data associated with this research will be made available at \url{https://github.com/arc-l/dcbs}. 

\subsection{Evaluation on globally dense instances}
\begin{figure}[h!]
    \centering
      \begin{overpic}               
        [width=\columnwidth]{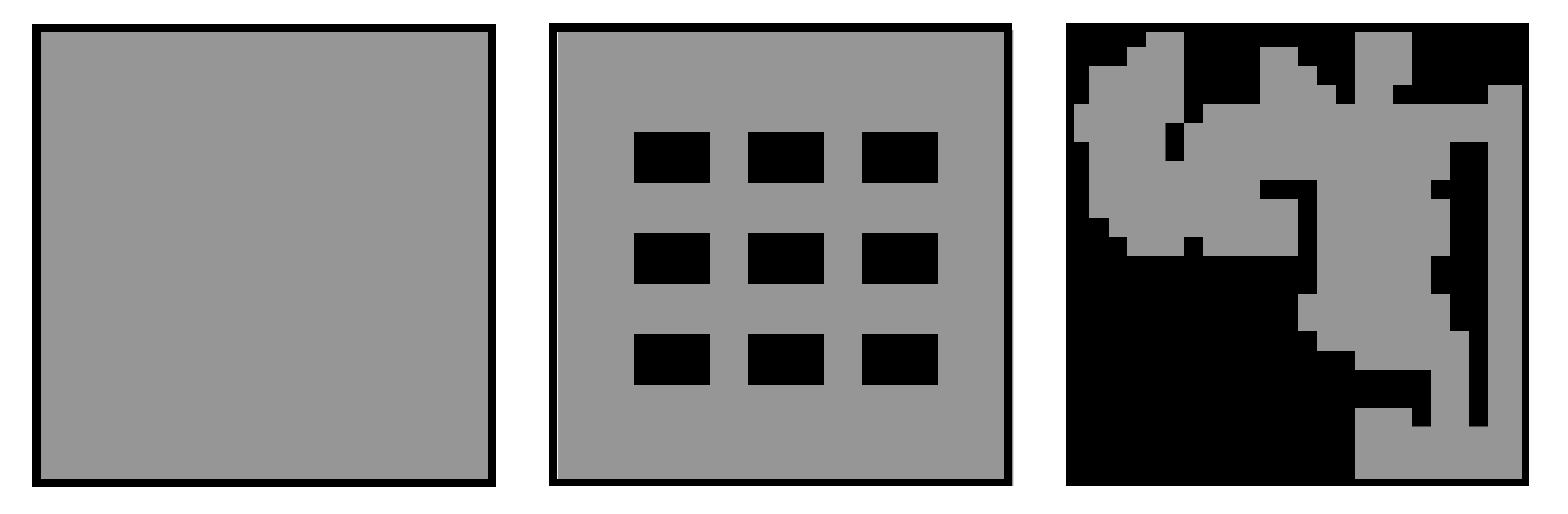}
             \small
             \put(15.5, -3) {(a)}
             \put(48.5, -3) {(b)}
             \put(81.5, -3) {(c)}
        \end{overpic}
    \vspace{1mm}
    \caption{The map used in the evaluation. (a) $20\times 20$ empty grid graph. (b) $24\times 18$ warehouse-like map. It has 360 non-blocked vertices. (c) $24\times 24$ ``lak103" game map adapted from DAO benchmarks \cite{stern2019mapf}. It has 293 non-blocked vertices.}
    \label{fig:maps_used}
    \vspace{2mm}
\end{figure}

\begin{figure}[h!]
    \centering
    \includegraphics[width=\linewidth]{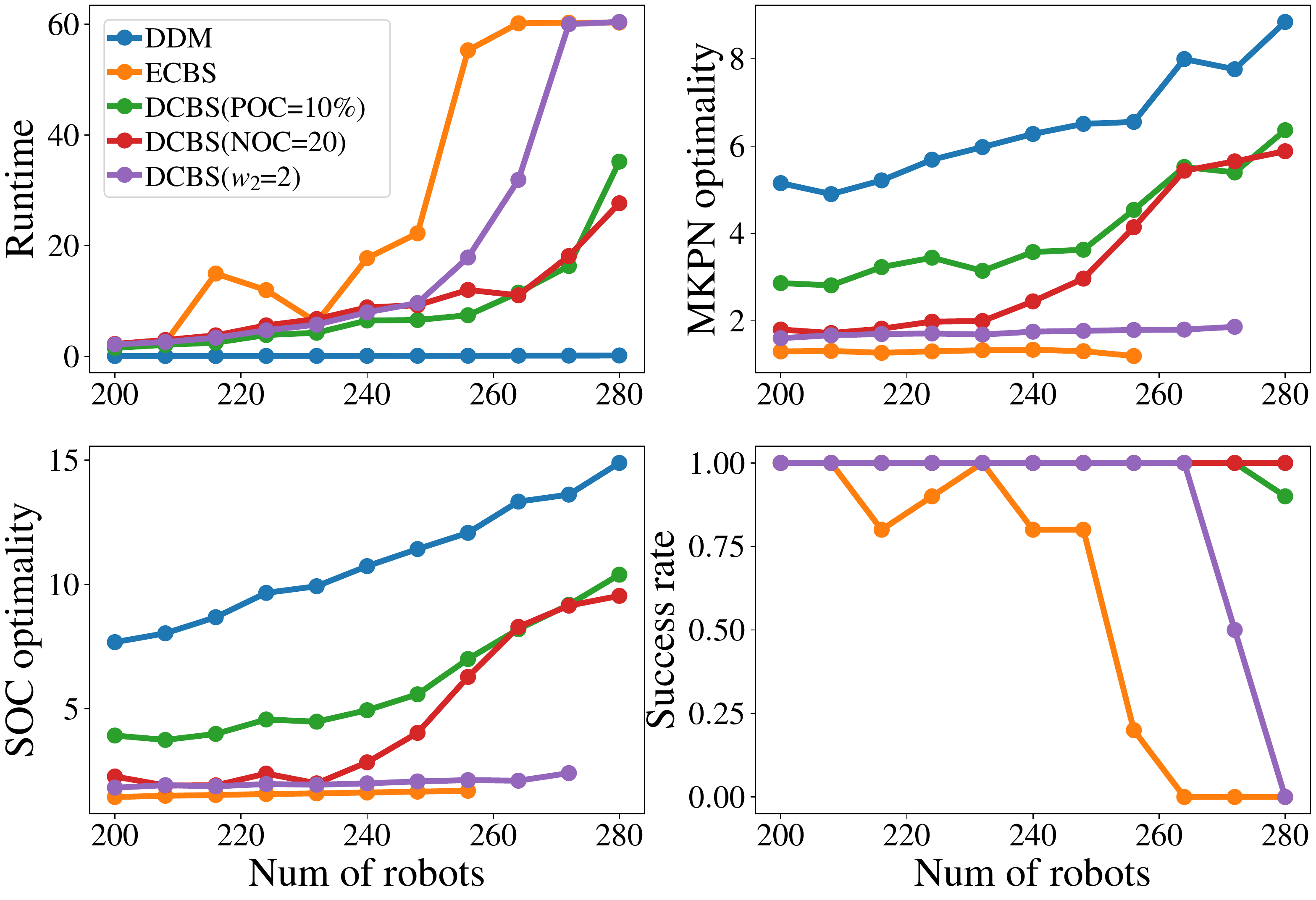}
    \caption{Performance (computation time, conservative makespan optimality ratio, conservative sum-of-cost optimality ratio, and success rate) on $20\times 20$ empty grid graph (Fig.~\ref{fig:maps_used}(a)) for DDM, ECBS, multiple \decbs variants. \decbs with $w_2 = 2$ scales much better than ECBS without losing much optimality guarantee. \decbs with POC- and NOC-based heuristics achieves an excellent balance between computation time and solution optimality.}
    \label{fig:20x20_data}
    \vspace{2mm}
\end{figure}

\begin{figure}[h!]
    \centering
    \includegraphics[width=\linewidth]{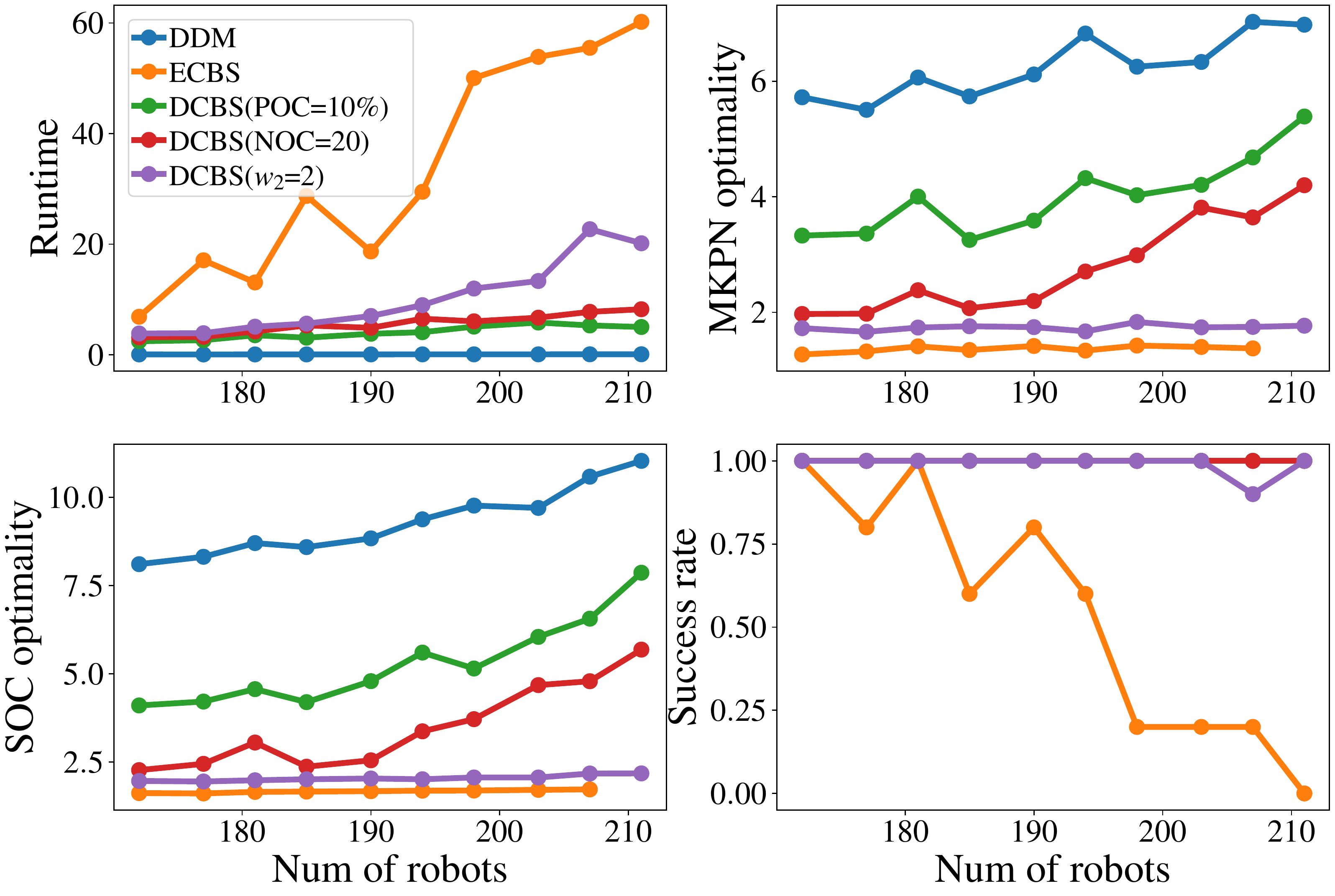}
    \caption{Performance (computation time, conservative makespan optimality ratio, conservative sum-of-cost optimality ratio, and success rate) on the warehouse map (Fig.~\ref{fig:maps_used}(b)) for DDM, ECBS, multiple \decbs variants. All \decbs variants achieve an excellent balance between computation time and solution optimality compared to DDM and ECBS; \decbs with $w_2 =2$ does especially well.}
    \label{fig:warehouse_data}
    \vspace{2mm}
\end{figure}

\begin{figure}[h!]
    \centering
    \includegraphics[width=\linewidth]{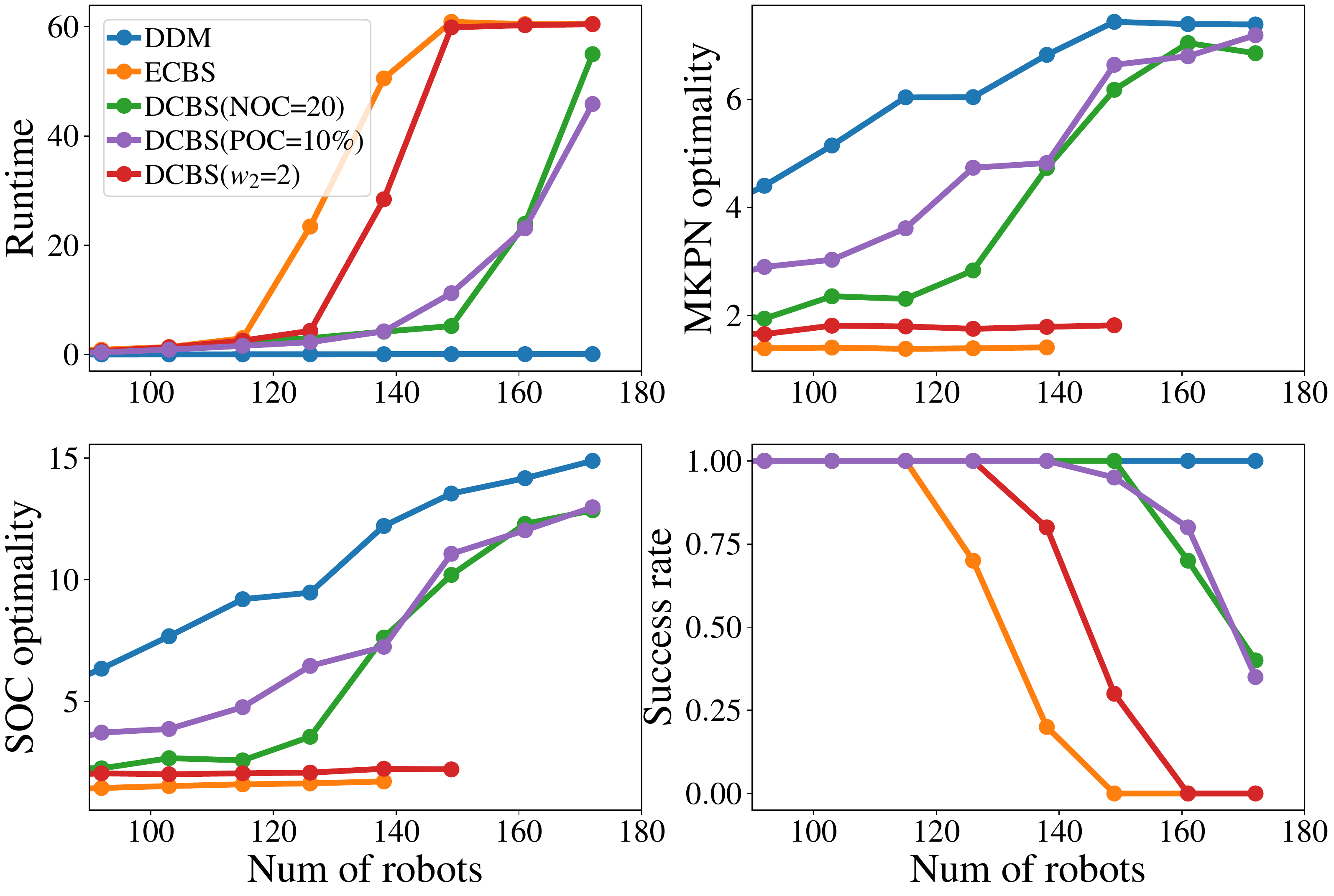}
    \caption{Performance (computation time, conservative makespan optimality ratio, conservative sum-of-cost optimality ratio, and success rate) on the DAO gamp map (Fig.~\ref{fig:maps_used}(c)) for DDM, ECBS, multiple \decbs variants. \decbs still does reasonably well in balancing solution computation speed and optimality.}
    \label{fig:lak103_data}
    \vspace{2mm}
\end{figure}

In this section, we evaluate \decbs on different maps with different high robot densities.
Here, the starts and goals are \emph{uniformly} randomly generated.
We evaluate the algorithms on three maps as shown in Fig.~\ref{fig:maps_used}.
The results are presented in Fig.~\ref{fig:20x20_data}-\ref{fig:lak103_data}.
Here, we tested three variants of \decbs.
They differ in the strategy used to start database conflict resolution.
\decbs(NOC=20) applies the database conflict resolution  when \noc of the high-level node drops below 20 and uses $w_2=\infty$.
\decbs(POC=$10\%$) applies the database conflict resolution when the ratio of the \noc of the current node to the \noc of the initial node  is less than $10\%$ and uses $w_2=\infty$. 
\decbs($w_2=2$) applies the database conflict resolution when it finds that the \noc enters stagnation for 100 iterations and uses $w_2=2$.
Here for \decbs($w_2=2$), we check the \makespan suboptimality. 

From the experimental data, we observe that the \makespan and \soc suboptimality ratio of \decbs variants are much better than \ddm.
When enabling the suboptimality checking mechanism, the suboptimality ratio of \decbs is around $1.x$, which is quite acceptable.
On the other hand, \decbs variants and \ddm are more scalable than \ecbs and thus yield a higher success rate.
On the empty grid and the warehouse map, the success rate of \decbs variants is almost always $100\%$, capable of tackling instances with robot density more than $60\%$-$70\%$.
On the  DAO map that is more complex and has some narrow passages, \decbs  is still able to solve more instances than \ecbs.
Despite the lower success rate, the suboptimality checking mechanism is essential to preserve the solution quality.

\subsection{Evaluation on locally dense instances}
In this section, we evaluate \secbs with DDM, ECBS, and \secbs in two classes of locally dense instances, named multi-robot rearrangement and Gaussian distributed \mpp instance.
To generate a Gaussian distributed \mpp instance,  for each point, we generate a 2D vector $(\lfloor x \rfloor,\lfloor y\rfloor)$ where $x,y\sim \mathcal{N}(0,\sigma^2)$ if point $(\lfloor x \rfloor,\lfloor y\rfloor)$ has not been used yet.
In the first class, the robots are randomly concentrated in the lower-left corner square area in start/goal configurations (e.g., the top row of Fig.~\ref{fig:dense_example}).
In the second class, the configurations are generated following a two-dimensional normal distribution with $\sigma=5$. 
In both classes, the graph size can be arbitrarily large (we set a sufficiently large boundary in the actual implementation).

\begin{figure}[h!]
    \centering
    \includegraphics[width=\linewidth]{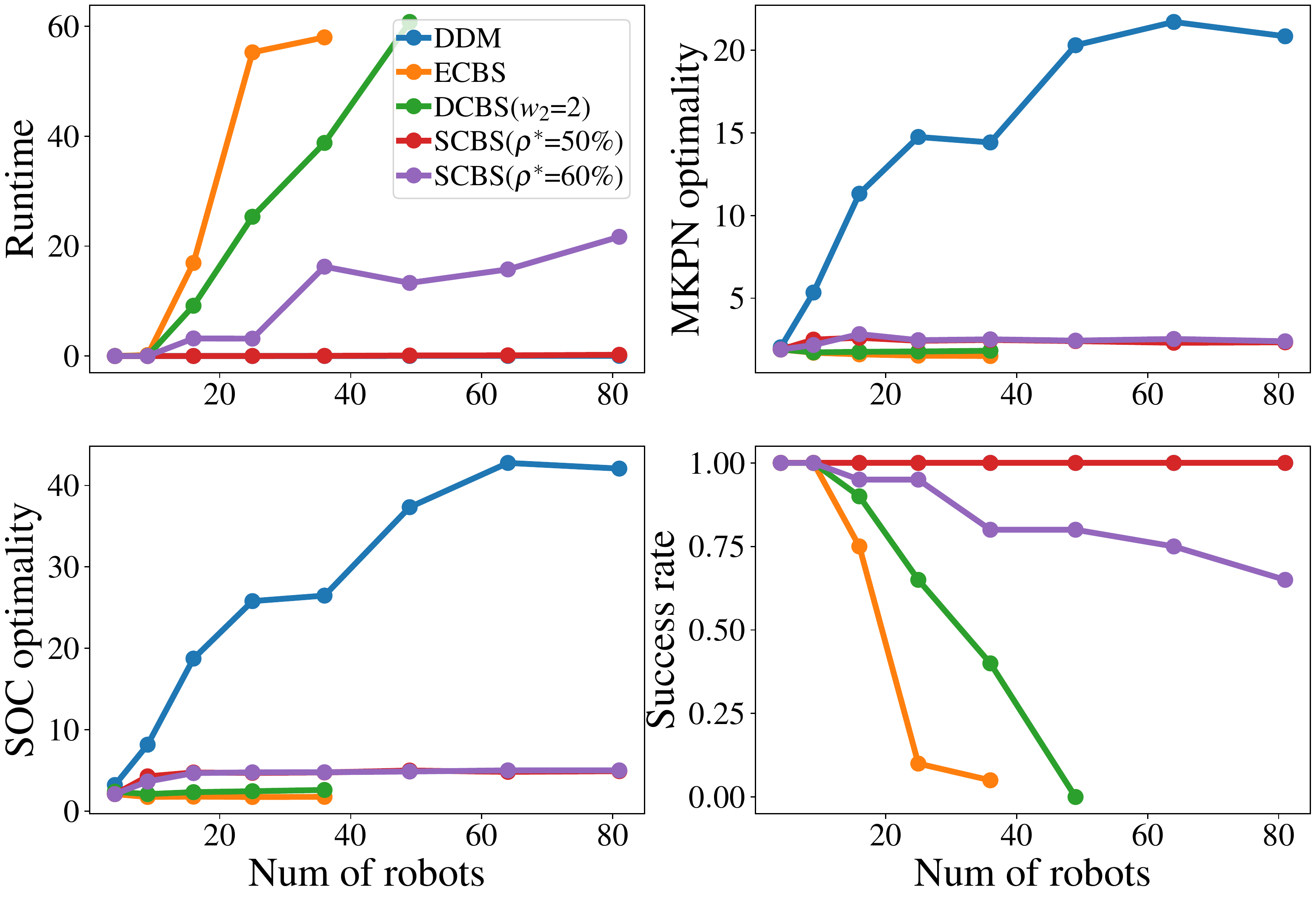}
    \caption{Performance (computation time, conservative makespan optimality ratio, conservative sum-of-cost optimality ratio, and success rate) on multi-robot rearrangement settings (e.g., the top row of Fig.~\ref{fig:dense_example}) for DDM, ECBS, \decbs, and \secbs. Whereas \decbs does better than DDM and ECBS, \secbs leaves all methods far behind in achieving an excellent balance between optimality and computational efficiency.}
    \label{fig:rearrangement}
    \vspace{2mm}
\end{figure}

\begin{figure}[h!]
    \centering
    \includegraphics[width=\linewidth]{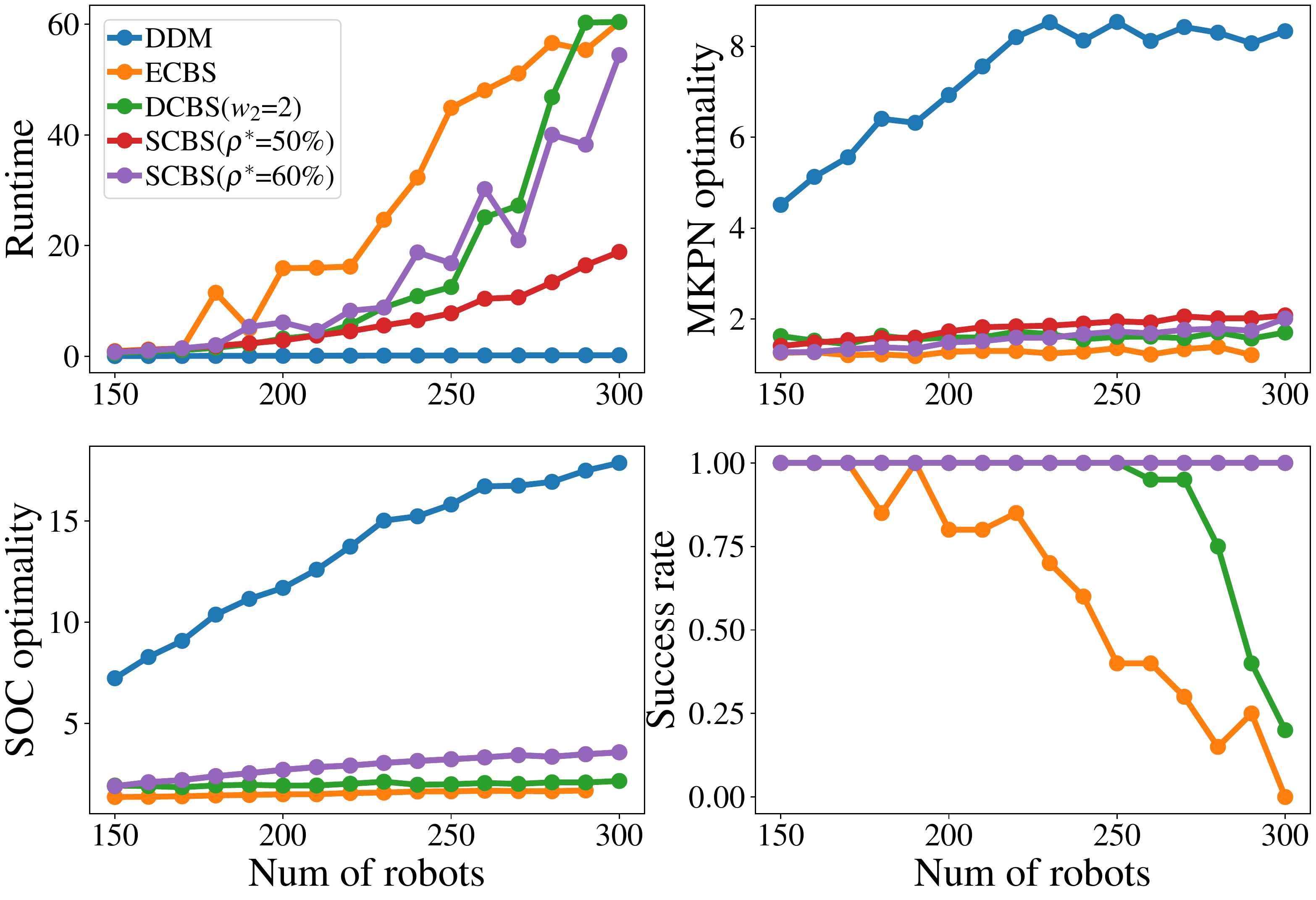}
    \caption{Performance (computation time, conservative makespan optimality ratio, conservative sum-of-cost optimality ratio, and success rate) on Gaussian distributed \mpp instances for DDM, ECBS, \decbs, and \secbs. Again, \secbs trades very nicely between scalability and solution optimality.}
    \label{fig:gauss}
    \vspace{2mm}
\end{figure}

The results are shown in Fig.~\ref{fig:rearrangement}-\ref{fig:gauss}.
In the first class (rearrangement), robots are so strongly-correlated that \ecbs struggles to solve instances with more than 36 robots.
\secbs($\rho^{*}=50\%$) yields $100\%$ success rate and is able to deal with 100+ robots.
The unlabeled \mpp only introduces small overheads to the solution, and the suboptimality ratio of \secbs is around $1.x$-$2.x$.

\vspace{-1mm}
\section{Conclusion and Discussions}\label{sec:conclusion}
In this paper, we present two novel heuristics-based algorithms for multi-robot path planning (\mpp) in dense and congested environments, with the goal to provide to quickly provide high-quality solutions for these problems. 
The first method, \decbs, incorporates a database-driven conflict resolution mechanism to resolve node conflicts in dense setups. Optimality protection rules are also instilled to maintain reasonable solution quality. 
Whereas \decbs addresses \emph{globally} dense scenarios, the second method, \secbs, tackles \emph{locally} dense settings by converting ultra-dense configurations into sparser ones through a greedy start-goal assignment and then solving an unlabeled \mpp. The sparsification step, while incurring some overhead, makes the overall problem significantly easier. 
Through extensive experiments, we show that our proposed methods achieve excellent performance in balancing success rate, running time, and solution quality.

Currently, \decbs only uses a fairly basic solution database, which is limiting the speed and flexibility of \decbs.  
In future work, we plan to significantly expand the solution database while keeping it sufficiently small for fast look-ups. Portions of the database may also be augmented using machine learning. We expect this to provide a sizable performance boost for \decbs. 

There are also many open questions that should be investigated further.
For example, as of now, the way we trigger the conflict resolution mechanism is somewhat rigid. Can we devise a better approach, e.g., using a data-driven method, to figure out the optimal time to trigger conflict resolution? 
As another example, there is still a lack of understanding of the exact relationship between time complexity and robot density and distribution. Can we establish a deeper, or better yet, quantitative, relationship between the two?

\bibliographystyle{formatting/IEEEtran}


\end{document}